\documentclass[lettersize,journal]{IEEEtran}
\usepackage{amsmath,amsfonts}
\usepackage{algorithmic}
\usepackage{array}
\usepackage{textcomp}
\usepackage{stfloats}
\usepackage{url}
\usepackage{verbatim}
\usepackage{graphicx}
\usepackage{hyperref}
\def\BibTeX{{\rm B\kern-.05em{\sc i\kern-.025em b}\kern-.08em
    T\kern-.1667em\lower.7ex\hbox{E}\kern-.125emX}}
\usepackage{balance}
\usepackage{multirow}
\usepackage{booktabs}
\usepackage{marvosym}
\begin{document}
\title{EgoPlan-Bench2: A Benchmark for \\Multimodal Large Language Model Planning\\ in Real-World Scenarios}

\author{\textbf{Lu Qiu$^{1,2,*}$, Yi Chen$^{1,2,*}$, Yuying Ge$^{2,}$\textsuperscript{\Letter}, Yixiao Ge$^2$,  Ying Shan$^2$, Xihui Liu$^{1,}$\textsuperscript{\Letter}} 

$^1$The University of Hong Kong, $^2$ARC Lab, Tencent PCG

\url{https://qiulu66.github.io/egoplanbench2/}
}



\maketitle

\begin{abstract}
The advent of Multimodal Large Language Models (MLLMs), leveraging the power of Large Language Models, has recently demonstrated superior multimodal understanding and reasoning abilities, heralding a new era for artificial general intelligence (AGI). However, achieving AGI necessitates more than just comprehension and reasoning. A crucial capability required is effective planning in diverse scenarios, which involves making reasonable decisions based on complex environments to solve real-world problems. Despite its importance, the planning abilities of current MLLMs in varied scenarios remain underexplored, leaving a significant gap in our understanding of their full potential.
In this paper, we introduce EgoPlan-Bench2, a rigorous and comprehensive benchmark designed to \textit{assess the planning capabilities of MLLMs across a wide range of real-world scenarios}. 
EgoPlan-Bench2 encompasses everyday tasks spanning 4 major domains and 24 detailed scenarios, closely aligned with human daily life. 
EgoPlan-Bench2 is constructed through a semi-automatic process utilizing egocentric videos, complemented by manual verification. Grounded in a first-person perspective, it mirrors the way humans approach problem-solving in everyday life.
We evaluate 21 competitive MLLMs and provide an in-depth analysis of their limitations, revealing that they face significant challenges in real-world planning. 
To further improve the planning proficiency of current MLLMs, we propose a training-free approach using multimodal Chain-of-Thought (CoT) prompting through investigating the effectiveness of various multimodal prompts in complex planning. Our approach enhances the performance of GPT-4V by 10.24\% on EgoPlan-Bench2 without additional training.
Our work not only sheds light on the current limitations of MLLMs in planning, but also provides insights for future enhancements in this critical area. 
We have made data and code available at \url{https://qiulu66.github.io/egoplanbench2/}.
\end{abstract}


\begin{figure*}[ht]
    \centering
    \includegraphics[width=1\textwidth]{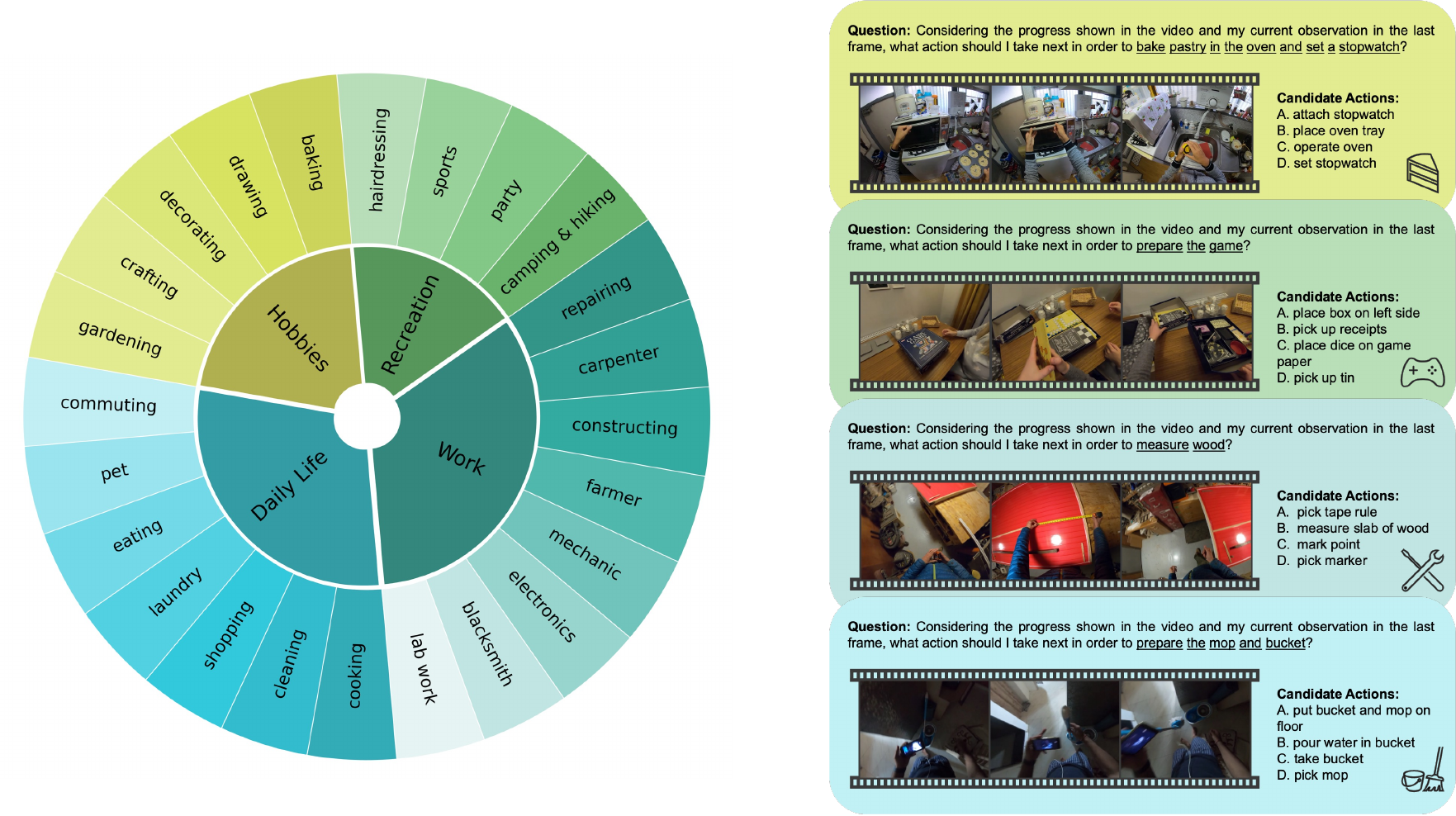}
    \caption{Left: EgoPlan-Bench2 encompasses planning tasks spanning four major domains and 24 detailed scenarios for evaluating the planning capabilities of MLLMs in \textbf{diverse real-world contexts}. Right: Examples of our multiple-choice question-answer pairs, where a partial video showing historical task progress, a current observation image, and a task goal expressed in language are given for a model to select the most appropriate action.}
    \label{fig:teaser}
\end{figure*}

\section{Introduction}
The rapid development of Multimodal Large Language Models (MLLMs)~\cite{team2023gemini,GPT4v,zhu2023minigpt,vlmsurvey} has demonstrated remarkable comprehension and generalization capacities, opening new possibilities for achieving the ultimate goal of artificial general intelligence (AGI)~\cite{y2023artificial,morris2023levels}, which aims to match or surpass human performance in most tasks. By plugging efficient visual encoders into pretrained Large Language Models (LLMs)~\cite{chowdhery2023palm,touvron2023llama,chiang2023vicuna,kenton2019bert} and learning alignments between vision and language~\cite{x2vlm}, MLLMs have 
excelled in various multimodal tasks such as image captioning~\cite{chen2015microsoft,li2024multimodal}, visual question answering~\cite{li2024multimodal,li2024mvbench,fu2024video,li2024seed,he2024mmworld,zhang2024q}, mathematical reasoning~\cite{lu2023mathvista,zhang2024mathverse}, crossing-modality grounding~\cite{peng2023kosmos}, etc. However, achieving AGI requires more than just advanced comprehension and reasoning. A crucial milestone is attaining human-level task \textit{planning} capabilities, which involve making informed decisions in complex environments. This capability is essential for developing a versatile intelligent assistant that can assist humans in tackling a wide array of real-world challenges in daily life.

While the comprehension capabilities of MLLMs have been extensively evaluated in previous benchmarks~\cite{liu2023mmbench, li2023seed, li2024seed, li2024seedplus, unk-vqa}, the evaluation of the planning abilities of MLLMs in various scenarios remains underexplored. A comprehensive benchmark specifically designed to assess the planning capabilities of MLLMs across diverse real-world scenarios is highly demanded to uncover the potential of MLLMs in serving as versatile assistants in the real world. Previous egocentric video question answering (QA) benchmarks~\cite{fan2019egovqa,jia2022egotaskqa} also evaluate model performance in everyday life, but they primarily assess comprehension rather than planning, where a model answers questions based on the spatial and temporal understanding of the entire video. Most relevant to addressing this issue is EgoPlan-Bench~\cite{chen2023egoplan}, which evaluates the planning abilities of MLLMs from an egocentric perspective. However, it is constrained to single kitchen scenarios, lacking a comprehensive evaluation across a variety of real-world contexts.

In this paper, we introduce \textbf{EgoPlan-Bench2}, a benchmark designed to rigorously assess the planning capabilities of MLLMs across a broad range of daily scenarios. EgoPlan-Bench2 is founded on three principal tenets: \textbf{a) Rich and diverse real-world scenarios.} It includes 1,321 high-quality multiple-choice QA pairs sourced from 1,113 videos, covering 4 major life domains: Work, Daily life, Hobbies and Recreation. These domains are further subdivided into 24 detailed scenarios, ranging from everyday household tasks to specialized activities such as laboratory work, blacksmith and mechanical repairs as shown in Fig.~\ref{fig:teaser}. In addition to scenario settings, EgoPlan-Bench2 features 284 distinct verbs in task goals and 434 in options, along with 742 and 1,113 unique objects respectively. The duration of task progress videos varies from a few seconds to five minutes. The rich scene setups, coupled with a variety of actions, objects and video lengths, ensure a comprehensive evaluation of MLLMs' planning capabilities across various contexts. \textbf{b) Egocentric perspective.} We choose Ego4D~\cite{grauman2022ego4d} as the video source, because it provides a vast array of first-person perspective videos that capture realistic human interactions with objects and environments. This egocentric perspective is crucial for evaluating planning capabilities in a manner that closely mirrors real-world human experiences, offering a more authentic assessment of how MLLMs can assist in everyday tasks. \textbf{c) Planning capability evaluation.} EgoPlan-Bench2 is specifically aimed at evaluating MLLMs’ planning abilities, where a model must track long-term task progress, comprehend the current state of the environment, and leverage both general and domain-specific knowledge to plan the next action, in order to correctly answer the questions as shown in Fig.~\ref{fig:teaser}.

The evaluation of task planning can either require the model to predict a sequence of actions or predict the next action. EgoPlan-Bench2 adopts a next action prediction evaluation protocol for the following two reasons:
\textbf{a) Dynamic decision-making simulation.} Humans adjust their decisions based on real-time observations when completing long-term tasks. Requiring the model to generate all action sequences at once does not accurately simulate this dynamic process. By focusing on next action prediction, EgoPlan-Bench2 allows the model to dynamically respond to changes and new observations, closely aligning with human decision-making processes.
\textbf{b) Foundation for sequence prediction.} Successfully predicting a single action lays the groundwork for accurate sequence prediction. In multi-step tasks, cumulative errors can lead to a higher failure rate. Therefore, it is logical to prioritize mastering single-action predictions before advancing to more complex action sequence predictions.

\begin{figure*}[ht]
    \centering
    \includegraphics[width=0.9\textwidth]{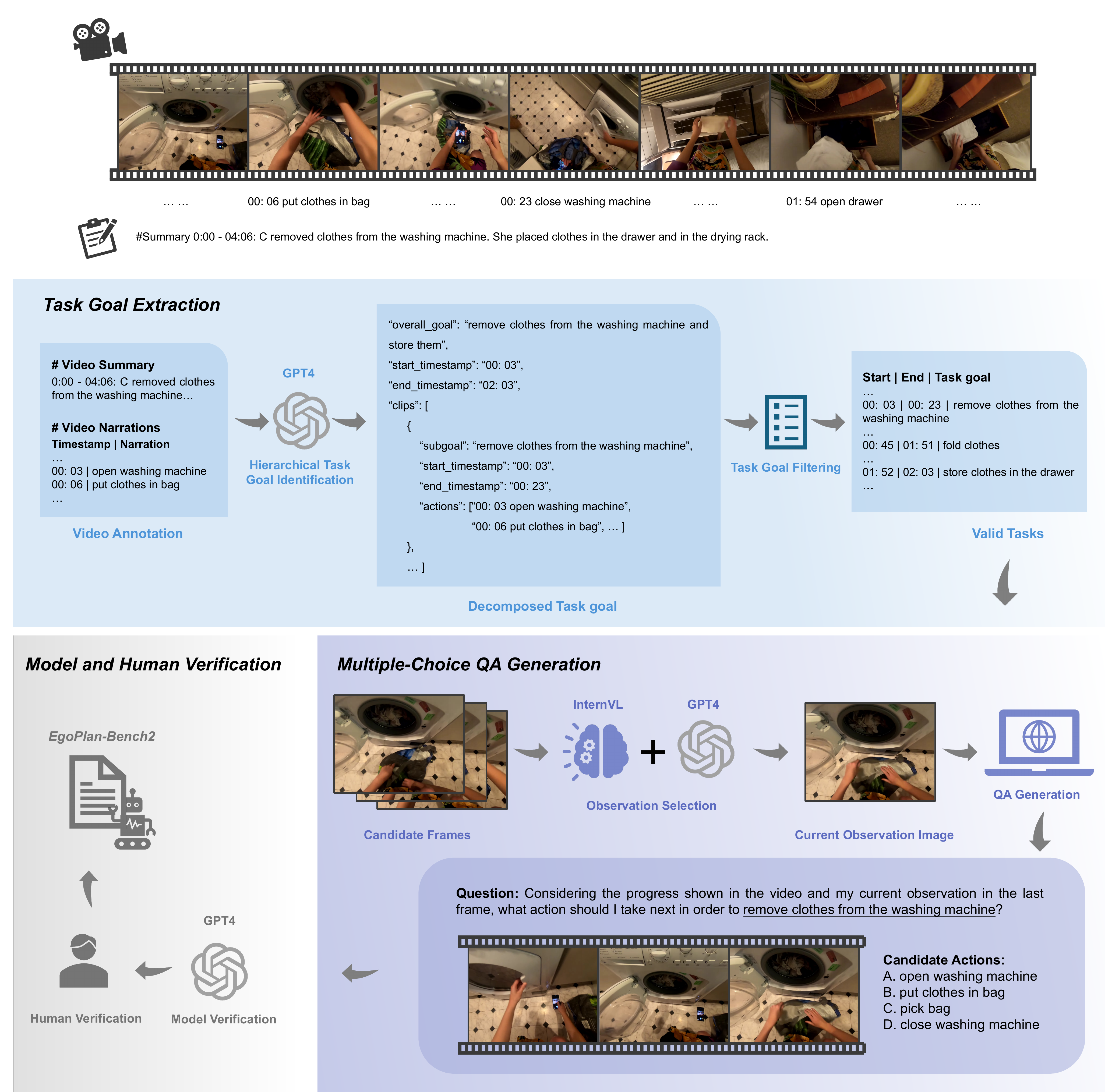}
    \caption{The overview of the semi-automatic dataset construction pipeline for EgoPlan-Bench2. \textbf{Stage I: Task Goal Extraction}, where task goals are summarized from video narrations by GPT-4 with a hierarchical extraction and decomposition strategy, and are further filtered to eliminate overly complex tasks. \textbf{Stage II: Multiple-choice QA Generation}, where multiple-choice questions are generated based on the task goals and corresponding action sequences using predefined templates. Foundation models are utilized to select an appropriate image as the visual observation (\textit{i.e.}, the end of the video showing task progress).  \textbf{Stage III: Model and Human Verification}, where model verification is conducted to reinforce the multimodal evaluation capability, and human annotators are employed to guarantee the reliability and objectivity of EgoPlan-Bench2.}
    \label{fig:pipeline}
\end{figure*}

To develop EgoPlan-Bench2, we design a semi-automatic dataset construction pipeline with three stages, as illustrated in Fig.~\ref{fig:pipeline}, to generate multiple-choice QA pairs based on egocentric videos in Ego4D. In \textbf{Stage I: Task Goal Extraction}, we employ a hierarchical task goal extraction and decomposition strategy, which utilizes GPT-4 to summarize task goals based on video narrations. The extracted task goals, along with their corresponding start and end timestamps and annotated action sequences, are further filtered to eliminate overly complex tasks that involve an excessive number of actions. In \textbf{Stage II: Multiple-choice QA Generation}, multiple-choice QA pairs are generated based on these task goals and corresponding action sequences using pre-defined templates. For each action designated as the groundtruth answer, the video segments occurring before its timestamps are selected to represent the historical task progress, and other three actions are randomly chosen as the distractor choices. To determine an appropriate image as the visual observation (\textit{i.e.}, the end of the video showing task progress), we utilize InternVL-1.5 and GPT-4 to ensure that the object involved in the groundtruth action is clearly depicted in this image. Additionally, we verify that the model cannot arrive at the correct answers based solely on this image, without taking the task progress into account. \textbf{Stage III: Model and Human Verification} focuses on reinforcing the multimodal evaluation capability and ensuring the reliability and objectivity of EgoPlan-Bench2. During model verification, questions that can be answered using only text input are removed. Rigorous quality control is then conducted through human verification, where a question is considered valid only if it can be correctly answered by human annotators.

We evaluate 21 competitive multimodal large language models, encompassing both proprietary and open-source models on EgoPlan-Bench2. The evaluation results demonstrate that our benchmark presents significant challenges for current MLLMs, revealing a substantial gap in their ability to achieve human-level task planning capabilities. Using GPT-4V, which achieves the best performance, as a case study, we analyze the underlying reasons for its shortcomings in real-world task planning. This detailed analysis provides valuable insights into areas for improvement that future research should prioritize.

To improve the planning proficiency of the most advanced GPT-4V, we propose a training-free approach that leverages multimodal Chain-of-Thought (CoT) prompting. We investigate the effectiveness of two primary categories of multimodal prompts: those focused on task progress and those centered on the current observation state. Our findings reveal that the most critical aspect of historical task progress is a precise and succinct sequence of actions with a clear temporal structure, rather than merely scene descriptions or object movements. Additionally, in terms of the current observation state, visual prompts such as bounding boxes that emphasize the interactions between objects and humans significantly improve the model’s planning accuracy. Through combining these effective multimodal prompts via CoT and a multi-iteration decision strategy, we enhance the performance of GPT-4V by 10.24\% on EgoPlan-Bench2 without additional training.

Our contributions can be summarized as four-fold:
\begin{itemize}
    \item We introduce EgoPlan-Bench2, a MLLM benchmark that provides a comprehensive assessment of task planning across various real-world scenarios, featuring 1,321 multiple-choice QA pairs spanning 4 primary domains and 24 fine-grained scenarios.
    \item We design an automated pipeline tailored for the unique challenges of noisy and uneven-quality egocentric videos, incorporating hierarchical task goal extraction and QA generation. We reinforce the dataset by a model and human verification phase to ensure quality and reliability.
    \item We assess a broad spectrum of MLLMs on EgoPlan-Bench2 and observe that existing MLLMs still face substantial challenges in planning tasks. Using the best performer GPT-4V as a case study, we analyze its performance pitfalls to guide future improvements.
    \item To improve the planning capabilities of MLLMs, we propose a training-free approach using multimodal CoT prompting through identifying effective multimodal prompts related to historical task progress and current observation state. Our approach achieves a significant performance improvement for GPT-4V.
\end{itemize}

\begin{figure*}[ht]
    \centering
    \includegraphics[width=1\textwidth]{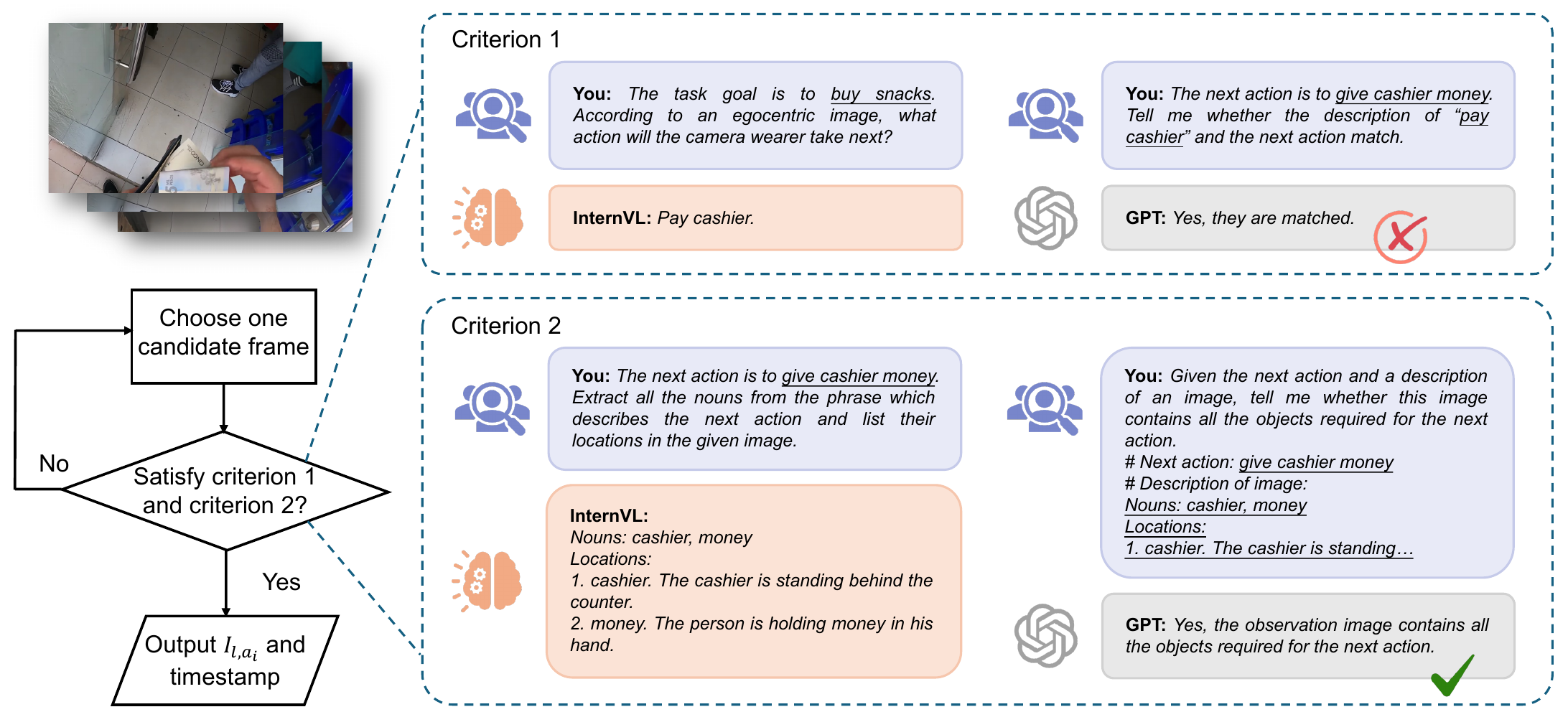}
    \caption{The pipeline of the adaptive observation selection method. Several frames around the timestamp of the groundtruth action are cropped as candidate frames. GPT-4 and InternVL-1.5 are then employed to verify whether each candidate frame is qualified. In this example, the selected candidate frame contains all objects necessary for the next action, fulfilling the second criterion. However, since InternVL-1.5 can correctly predict the upcoming action without historical task progress information, this frame fails to meet the first criterion and should therefore be discarded.}
    \label{fig:observation_select}
\end{figure*}

\section{Related Work}
\subsection{Advancements in Multimodal Large Language Models}
Building upon the impressive achievements of LLMs, MLLMs have also experienced a revolutionary transformation. MLLMs typically consists of an image encoder (e.g., CLIP~\cite{radford2021learning}) to extract visual information, a language model (e.g., LLaMA~\cite{touvron2023llama}, Vicuna~\cite{chiang2023vicuna}) to decode multimodal or text sequence and a trainable align module (e.g., Q-Former~\cite{li2023blip}, gated cross-attention layer~\cite{alayrac2022flamingo}) to integrate visual features into the language embedding space. Groundbreaking models like BLIP-2~\cite{li2023blip}, LLaVA~\cite{liu2024visual}, Flamingo~\cite{alayrac2022flamingo} and PaLM-E~\cite{driess2023palm} have made early attempt to integrate LLMs into vision-language pre-training and have demonstrated remarkable multimodal understanding and reasoning capabilities. Recent research interest has increasingly shifted towards multimodal understanding and generation that incorporates videos as visual signals~\cite{li2023videochat,maaz2023video,ge2023planting,ge2023making,ge2024seed}. These methods (e.g., VideoChat~\cite{li2023videochat}, VideoChatGPT~\cite{maaz2023video}, Valley~\cite{luo2023valley}) try to enhance MLLMs' instruction-following capabilities by generating video instruction-tuning data. Video-LLaMA~\cite{zhang2023video} encodes individual frames through a ViT~\cite{dosovitskiy2020image} and an image Q-Former and then apply temporal process through a video Q-Former. VideoChat2~\cite{li2024mvbench} encodes video frames through a video transformer, and Q-former is employed to compress video tokens. Expect these models mentioned above, a considerable number of video-based MLLMs~\cite{chen2024sharegpt4video,zhang2024llavanextvideo,zhang2024longva,cheng2024videollama} have been proposed, demonstrating notable generalization and reasoning abilities across a wide range of tasks.

\subsection{Benchmarking Multimodal Large Language Models}
To guide the potential future development of MLLMs, numerous benchmarks have been proposed to assess model performance across various aspects and tasks. Alongside advancements in comprehensive benchmarks for image MLLMs~\cite{liu2023mmbench,xu2023lvlm,ye2023mplug,yin2024lamm,yue2024mmmu,ying2024mmt}, significant efforts have been made to creating benchmarks for video MLLMs~\cite{patraucean2024perception,maaz2023video,fu2024video,li2024mvbench,he2024mmworld}. For example, Video-MME~\cite{fu2024video} is build upon videos collected from YouTube, aiming to evaluate models' capacities in 12 task types involving temporal perception, spatial perception, OCR, object recognition, etc. MVBench~\cite{li2024mvbench} focuses on temporally-sensitive videos and conducts comprehensive evaluations of MLLMs’ temporal understanding. MMWorld~\cite{he2024mmworld} is a benchmark characterized by its multi-discipline nature, evaluating models on multi-faceted tasks such as explanation and counterfactual thinking. MLVU~\cite{zhou2024mlvu} is a long video understanding benchmark which is built upon various video genres, including movies, cartoons, game videos, etc. The diagnostic benchmark dataset, WorldQA~\cite{zhang2024worldqa}, challenges machines to answer questions about a video by employing multimodal data (auditory and visual) and world knowledge. Despite the advancement, videos in these benchmarks are not recorded from a first-person perspective and fail to simulate the realistic visual input that a model would receive during planning tasks. In addition, these benchmarks mainly focus on MLLMs' comprehension capabilities instead of planning capabilities.

\subsection{Egocentric Video Datasets}
Numerous egocentric datasets containing daily life activity have been developed over the past years, including Ego4D~\cite{grauman2022ego4d}, Epic-Kitchens~\cite{epickitchen}, UT Ego~\cite{lee2012discovering,su2016detecting}, Activities of Daily Living (ADL)~\cite{pirsiavash2012detecting}, Disney dataset~\cite{fathi2012social}, Charades-Ego~\cite{sigurdsson2018charades}, etc. Among these egocentric datasets, we select Ego4D as the video source for EgoPlan-Bench2 due to its massive scale, encompassing hundreds of indoor and outdoor environments, a much wider demographic and an exceptionally rich variety of tasks and scene types. There are also some existing egocentric QA benchmarks like EgoThink~\cite{cheng2023can}, EgoVQA~\cite{fan2019egovqa} and Egotaskqa~\cite{jia2022egotaskqa}, but they mainly focus on evaluating model's reasoning and comprehension abilities of activities, human-object interactions and environments instead of task planning capacity. Most relevant to our work are EgoPlan-Bench~\cite{chen2023egoplan} and VidEgoThink~\cite{cheng2024videgothink}. EgoPlan-Bench is also a planning benchmark that utilizes kitchen-related videos from Epic-Kitchens~\cite{epickitchen} and Ego4D datasets, but it focuses exclusively on cooking scenario and lacks evaluation across a wider variety of real task types. VidEgoThink evaluates the capabilities for different functions of MLLMs in Embodied AI from four dimensions: video QA, hierarchy planning, visual grounding and reward modeling. In the most relevant dimension of hierarchy planning, VidEgoThink includes only 9 scenes sourced from 76 videos, suffering from limited task types and scenario diversity.

\section{Constructing EgoPlan-Bench2}
To simulate how MLLMs function as versatile AI assistants in managing complex tasks, the proposed EgoPlan-Bench2 is founded on three essential design principles: a) diverse scenarios reflective of real-world human life, b) an egocentric perspective, and c) a focus on evaluating planning tasks.

Our methodology begins with the collection of a comprehensive set of egocentric videos that cover 4 fundamental domains of human life, contributing to the properties of egocentric perspective and diverse scenarios of EgoPlan-Bench2. In terms of the last principle, we design a semi-automatic dataset construction pipeline to generate high-quality QA pairs focusing on planning tasks. Finally, we provide the detailed data statistics of EgoPlan-Bench2.

\subsection{Egocentric Video Source}
\label{sec:video_source}
Ego4D is a massive-scale video dataset that captures human activities from a first-person perspective across a wide range of scenarios (household, outdoor, workplace, leisure, etc.), making it an exemplary data source for planning tasks. Because of its unprecedented scale and diversity, Ego4D is chosen as the video source for our benchmark. In this paper, we utilize the updated version (\href{https://ego4d-data.org/docs/updates/}{https://ego4d-data.org/docs/updates/}) containing 3,900 hours of 9,611 egocentric videos. The selected videos are categorized into 24 scenarios which are important and ubiquitous in real-world planning tasks and closely reflect human daily life. As shown in Fig.~\ref{fig:teaser}, we summarize them into 4 major domains to simplify experiment analysis:
\begin{itemize}
    \item Work (8 scenarios): lab work, blacksmith, electronics, mechanic, farmer, constructing, carpenter, repairing.
    \item Daily life (7 scenarios): commuting, pet, eating, laundry, shopping, cleaning, cooking.
    \item Hobbies (5 scenarios): baking, drawing, decorating, crafting, gardening.
    \item Recreation (4 scenarios): camping \& hiking, party, sports, hairdressing.
\end{itemize}

Ego4D has densely timestamped annotations indicating specific actions and their occurrence time, represented by short sentences in the format of ``\textit{\#C C does something}". In order to remove invalid annotations and reduce noise, we adopt the criteria following the EgoVLP framework~\cite{lin2022egocentric}:
\begin{itemize}
    \item Filter narrations with unsure tags, e.g., ``\textit{\#C C washes \#unsure in sink}".
    \item Remove narrations less than three words. Because such narrations generally do not include effective interactions with environment or objects, e.g., ``\textit{\#C C speaks}", ``\textit{\#C C looks}".
    \item Exclude narrations annotated with ``\textit{\#O}", which indicate actions performed by individuals other than the camera wearer.
\end{itemize}
After narration filtering, we convert raw narrations to verb-object phrases like ``\textit{close washing machine}" with GPT-4 to better represent actions and uniform format. The raw action timestamps in Ego4D indicate when actions occur but not their durations. Following EgoVLP~\cite{lin2022egocentric}, we calculate start and end timestamps for each action. For an action with occurrence timestamp $t_i$, the start and end timestamps can be calculated as:
\begin{equation}
    [{t_i}^{start}, {t_i}^{end}] = [t_i - \beta_i/2\alpha, t_i + \beta_i/2\alpha],
\end{equation}
where $\beta_i$ is the average temporal distance between pairs of consecutive
narrations in this video, and $\alpha$ is the scale factor calculated from the whole dataset ($\alpha = 4.9$).

\subsection{Construction Pipeline}
\label{sec:pipeline}
To generate multiple-choice questions focusing on evaluating planning capabilities of MLLMs, we design a goal-oriented semi-automatic dataset construction pipeline starting with the task goal extraction, as illustrated in Fig.~\ref{fig:pipeline}. After obtaining valid task goals with unified format, we generate multiple-choice questions based on the goal-action pairs and finally confirm its validity via GPT-4 and human annotators.

\subsubsection{Stage I: Task Goal Extraction}
Egocentric videos in Ego4D exhibit significant variability. Some videos documenting simple tasks are relatively short, while others are much longer, capturing complex activities that may span a considerable duration and encompass multiple distinct tasks within a single video clip. Additionally, many of the activities recorded in these videos are purposeless or aimless, such as walking or engaging in casual conversations. This inherent variability and the prevalence of non-goal-oriented activities present substantial challenges in identifying task goals.

To address these problems, we design a \textbf{hierarchical task goal identification} strategy to extract task goals. Ego4D provides sub-segment summaries for a single video, for example, ``\textit{0:00 - 05:00 \#Summary C was at the kitchen, googled on phone and watched video on the phone; 05:00 - 09:00 \#Summary C was in the kitchen alone and mixed flour in a plate}". We use the time intervals provided in the summaries to divide each video into sub-segments and then process each sub-segment separately. GPT-4 takes the summary of the sub-segment along with the corresponding actions as input. GPT-4 extracts the overall task goal and decomposes it into sub-goals and action sequences. For instance, the overall goal ``\textit{remove clothes from washing machine and store them}" can be divided into sub-goals of ``\textit{remove clothes from washing machine}", ``\textit{fold clothes}" and ``\textit{store clothes in the drawer}". The sub-goal ``\textit{remove clothes from washing machine}" consists of actions such as ``\textit{open washing machine}" and ``\textit{put clothes in bag}". By employing this hierarchical approach, GPT-4 can effectively process video segments of varying complexity, arranging them into a structured framework. To avoid including purposeless activities which are not suitable for planning tasks, we require the process of extraction and decomposition of task goals to adhere to the following criteria:
\begin{itemize}
    \item The overall task goals must be purposeful, meaning that the camera wearer is engaged in a task with logical steps and a clear purpose. For example, planting flowers requires digging a hole, placing seeds, covering soil, etc.
    \item The sub-goals should also be purposeful.
    \item All actions must be directed towards achieving the corresponding sub-goal. It means any actions that are not relevant to the corresponding sub-goal should be removed.
\end{itemize}
Furthermore, we implement an additional verification step to ensure that the action sequences generated by GPT-4 are not fabricated and that the timestamps are accurate. Please refer to supplementary materials for detailed prompts.

Even the overall goal of a short segment can be simpler than just a sub-goal in a longer segment, leading to inconsistencies in the alignment of task hierarchies across various videos and their sub-segments. To mitigate substantial discrepancies in task complexity, we introduce the \textbf{task goal filtering} step to consolidate all the overall goals and sub-goals (collectively referred to as ``task goal" hereafter) by filtering them based on the number of actions. We retain these task goals with 4-20 actions to ensure a moderate level of task complexity.

\subsubsection{Stage II: Multiple-choice QA Generation}
In this section, we generate multiple-choice QA pairs, which simplifies the evaluation process compared to open-ended QA, in an automated manner from filtered task goals with corresponding action sequences. The QA pairs in EgoPlan-Bench2 incorporate multimodal inputs, including textual questions and visual information representing historical task progress and current observation state. For the textual input, we create goal-action pairs and transform them into multiple-choice QA format using predefined templates. For the visual input, video segment preceding the timestamp of the groundtruth action represent the historical task progress and current observation state. To ensure proper alignment with the corresponding question, we design an adaptive method for selecting the optimal current observation image and determining the appropriate video segment, as shown in Fig.~\ref{fig:observation_select}.

a. Create goal-action pairs and QA pairs. We utilize a predefined template as shown in the solid purple part in Fig.~\ref{fig:pipeline} to generate QA pairs. To be more specific, given a task goal $l$ with $N$ actions ${a_1, ..., a_i, ..., a_N}$, we can obtain $N$ corresponding goal-action pairs $[l, a_i], i=1,2,...,N$. The underlined words in the template are replaced with $l$, and action $a_i$ is the groundtruth answer of the corresponding QA pair. For candidate actions in the predefined template, we select three actions from different timestamps within the same task goal as negative choices, in order to assess models' ability of task progress-related temporal understanding. To mitigate semantic similarity among different choices within the same QA pair, we employ GPT-4 to categorize all actions from the same task goal based on their semantic content. Negative choices are then randomly selected from three distinct categories that differ from the ground truth.

b. Align visual input with question. Given a specific question defined on the goal-action pair $[l, a_i]$, the visual input consists of a video clip $H_{l,a_i}$ representing the historical task progress and an image $I_{l,a_i}$ representing the current observation scene. We combine them into a total video input $V_{l,a_i}$, where the final frame represents the current observation image. In order to cut the video input $V_{l,a_i}$ from the full video, we need to determine the start and end timestamps of $V_{l,a_i}$. The start timestamp of $V_{l,a_i}$ is set to the start timestamp of the first action $a_1$, which means the beginning of the whole task goal. However, it is challenging to determine the end timestamp of $V_{l,a_i}$, which shows a suitable and qualified current observation. The first challenge is that, an ideal visual input $V_{l,a_i}$ should encompass all completed historical actions, while excluding any frames related to the next action. It means the timestamp of $I_{l,a_i}$ must fall after the completion of the last action and before the start of the next action. Another challenge is that, $I_{l,a_i}$ should clearly represent the current observation state, capturing all manipulated objects without revealing any clues (e.g., hand-object interactions) about the groundtruth answer.

An intuitive mechanism to select $I_{l,a_i}$ is using the end timestamp of the last action $a_{i-1}$ or the start timestamp of the next action $a_{i}$. However, variations in action durations and prevalence of perspective shifting can result in selected frames being blurred, missing manipulated objects, or overlapping with the onset of the next action. To overcome this problem, we introduce an \textbf{adaptive observation selection method} that choose the optimal $I_{l,a_i}$ from multiple candidate frames cropped around the start timestamp of $a_{i}$, which is illustrated in Fig.~\ref{fig:observation_select}. To be more specific, we set 0.5 seconds preceding the timestamp of the groundtruth action $a_i$ as the baseline and select candidate frames at 0.25-second intervals, extracting five frames sequentially. For each candidate frame, we employ InternVL-1.5 and GPT-4 to verify whether it satisfies the following two key criteria:
\begin{itemize}
    \item \textbf{Prevent models from cheating with the clues from hand-object interaction.} When the subsequent action has already begun, it may be feasible to infer the next action solely from the current observation image. This possibility is contrary to our expectations that the evaluation should focus on comprehensive task understanding and planning. For this reason, we adopt InternVL-1.5 to determine the next action only based on a candidate frame and utilize GPT-4 to assess whether it aligns with the ground truth. A correct answer from InternVL-1.5 indicates that this frame is unqualified.
    \item \textbf{The manipulated objects are clearly visible.} Visual occlusion and rapid movement generally exhibit in the first-person perspective videos, and the object to manipulate in the next action is not always clearly visible, which hinders the subsequent task planning. For example, if the scissors are not visible in the observation image, the model cannot choose to pick up scissors. We also prompt InternVL-1.5 and judge the answer with GPT-4 in this phase. We provide detailed instructions to make answers of InternVL-1.5 more controllable, by requiring it to first list all the nouns in the next action and then locate the object represented by each noun.
\end{itemize}
Only the frame achieving both criteria can be chosen as the current observation image $I_{l,a_i}$. We can determine the end timestamp of $V_{l,a_i}$ and align the visual input with the question.

\subsubsection{Stage III: Model and Human Verification}
In the final verification phase, we exclude QA pairs that can be correctly answered using only textual information, ensuring the benchmark effectively assesses the models’ capabilities to process multimodal inputs. GPT-4 is employed with the CircularEval Strategy~\cite{liu2023mmbench} to minimize the impact of random guessing.

To further bolster the reliability and objectivity of EgoPlan-Bench2, we introduce a human verification step. After carefully reviewing the video segment $V_{l,a_i}$ and the associated questions, annotators select the most suitable answer for each QA pair and remove those that are unqualified, including any with unclear options or that with blurry and low-quality video inputs. A question is considered valid and can be retained if annotators can answer it consistent with the ground truth.

\begin{figure*}[ht]
    \centering
    \includegraphics[width=1\textwidth]{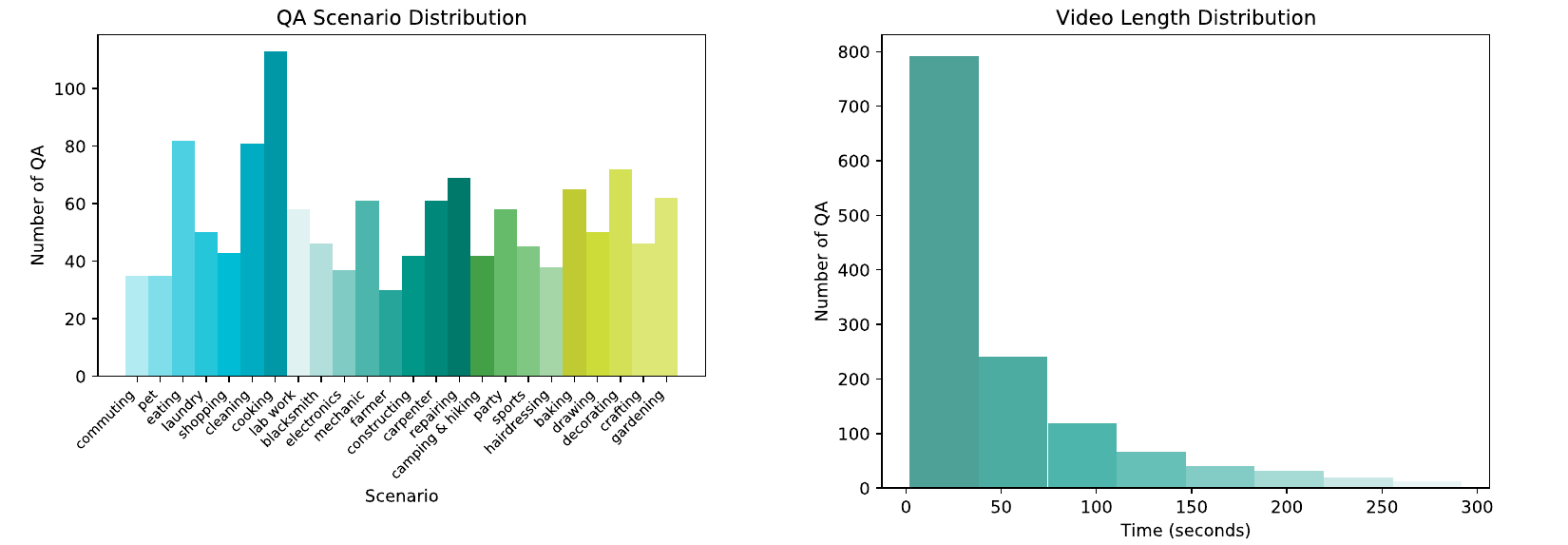}
    \caption{Left: Scenarios distribution of EgoPlan-Bench2, which covers 4 major domains and 24 fine-grained scenarios. Right: Video length distribution. Our benchmark has a full spectrum of video duration, ranging from a few seconds to five minutes.}
    \label{fig:data_statistic}
\end{figure*}

\begin{figure*}[ht]
    \centering
    \includegraphics[width=1\textwidth]{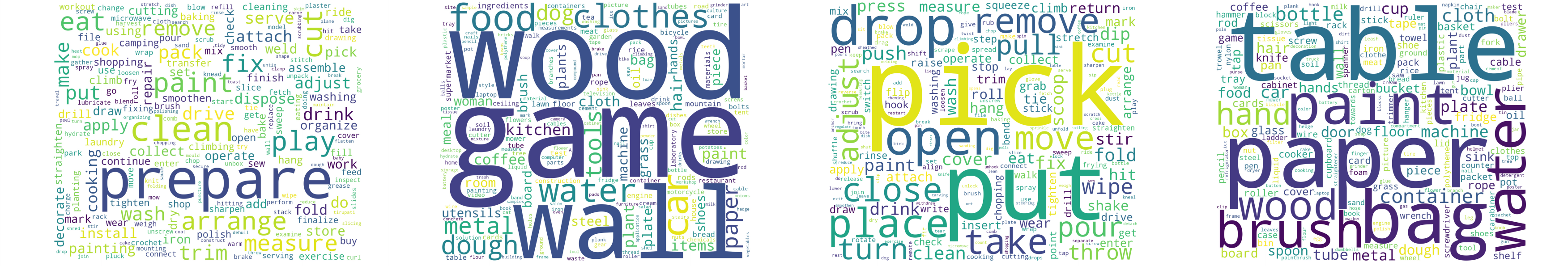}
    \caption{Word clouds of task goals and candidate options in EgoPlan-Bench2. From left to right: verbs in task goals, objects in task goals, verbs in candidate options, objects in candidate options.}
    \label{fig:word_cloud}
\end{figure*}

\subsection{EgoPlan-Bench2 Statistics}

EgoPlan-Bench2 is highly diverse in terms of scenarios, human-object interactions, and video durations, comprising 1,321 high-quality multiple-choice QA pairs derived from 1,113 videos. Firstly, it has an \textbf{extensive range of scenario} settings across 4 major life domains: Work (404 QA pairs), Daily Life (439 QA pairs), Recreation (183 QA pairs) and Hobbies (295 QA pairs). A more detailed breakdown of the 24 fine-grained scenarios and the corresponding number of QA pairs is provided in Fig.~\ref{fig:data_statistic}. Second, EgoPlan-Bench2 incorporates \textbf{comprehensive human-object interaction data}, manifested in both actions and task goals. We use NLTK~\cite{bird2009natural} to analyze the verbs and nouns in the task goals and candidate options of the QA pairs, identifying 284 distinct verbs and 742 unique objects in the task goals, and 434 verbs and 1,113 objects in the candidate options. The word clouds of these terms are displayed in Fig.~\ref{fig:word_cloud}. Third, \textbf{varied video durations} in EgoPlan-Bench2 range from a few seconds to five minutes. We consider 30 seconds as the threshold, classifying videos of 30 seconds or less as short videos (690 QA pairs) and those exceeding this duration as long videos (631 QA pairs).

\section{Experiments}

\begin{table*}
\begin{center}
\caption{Performance of 21 MLLMs on EgoPlan-Bench2.}
\label{tab:main_results}
\begin{tabular}{@{\extracolsep{4pt}}cccccccccc}
    \toprule
    \multirow{2.5}{*}{Model} & \multirow{2.5}{*}{Frames} & \multirow{2.5}{*}{LLM} & \multicolumn{4}{c}{Domain} & \multicolumn{2}{c}{Video Length} & \multirow{2.5}{*}{Total Acc}\\
    \cmidrule(lr){4-7} \cmidrule(lr){8-9}
    & & & Daily life & Work & Recreation & Hobbies & $\leq$30s & $>$30s & \\
    \midrule
    \multicolumn{10}{c}{\textit{Image MLLMs}} \\
    \midrule
    Yi-VL\cite{young2024yi} & 8 & Yi-6B  & 24.37 & 21.29 & 26.23 & 23.39 & 23.19 & 23.77 & 23.47\\
    MultiModal-GPT\cite{gong2023multimodal} & 8 & LLaMA-7B & 26.42 & 23.27 & 23.50 & 26.10 & 23.48 & 26.62 & 24.98\\
    LLaVA1.5\cite{liu2024improved} & 6 & LLaMA-7B & 29.61 & 21.04 & 27.32 & 24.07 & 23.77 & 27.26 & 25.44\\
    InternVL-1.5\cite{chen2024internvl,chen2024far} & 8 & InternLM2-Chat-1.8B & 28.02 & 24.75 & 23.50 & 24.41 & 23.91 & 27.42 & 25.59\\
    mPLUG-Owl-2\cite{ye2024mplug} & 8 & LLaMA2-7B & 27.79 & 24.75 & 24.04 & 25.42 & 25.36 & 26.31 & 25.81\\
    BLIP-2\cite{li2023blip} & 8 & Flan-T5-XL & 24.37 & 23.51 & 30.05 & 30.17 & 24.64 & 27.89 & 26.19\\
    InstructBLIP\cite{dai2023instructblip} & 8 & Flan-T5-XL & 27.33 & 23.02 & 26.23 & 29.49 & 25.51 & 27.26 & 26.34\\
    InstructBLIP Vicuna\cite{dai2023instructblip} & 8 & Vicuna-7B & 27.56 & 24.26 & 28.42 & 28.14 & 25.65 & 28.05 & 26.80\\
    DeepSeek-VL\cite{lu2024deepseek} & 6 & DeepSeek-LLM-7B & 32.12 & 24.75 & 26.23 & 29.83 & 28.55 & 28.53 & 28.54\\
    Qwen-VL-Chat\cite{bai2023qwen} & 8 & Qwen-7B & 32.57 & \underline{27.23} & 27.87 & 28.47 & 30.00 & 28.68 & 29.37\\
    InternVL-2\cite{chen2024internvl,chen2024far} & 8 & InternLM2.5-Chat-7B & \textbf{37.81} & 23.76 & \underline{31.69} & 28.14 & \underline{31.01} & \underline{29.95} & \underline{30.51}\\
    \toprule
    \multicolumn{10}{c}{\textit{Video MLLMs}} \\
    \midrule
    Video-LLaMA2\cite{cheng2024videollama} & 8 & Mistral-v0.2-Instruct-7B & 24.15 & 23.02 & 19.13 & 23.73 & 23.19 & 22.82 & 23.01\\
    LLaVA-NeXT-Video\cite{zhang2024llavanextvideo} & 16 & Vicuna1.5-7B & 26.42 & 19.55 & 24.59 & 23.05 & 22.61 & 24.09 & 23.32\\
    Video-ChatGPT\cite{maaz2023video} & 100 & LLaMA-7B & 24.15 & 22.77 & 24.59 & 24.07 & 23.33 & 24.25 & 23.77\\
    Video-LLaVA\cite{lin2023video} & 8 & Vicuna1.5-7B & 27.11 & 22.52 & 27.87 & 24.75 & 25.22 & 25.36 & 25.28\\
    ShareGPT4Video\cite{chen2024sharegpt4video} & 16 & LLaMA3-Instruct-8B & 25.51 & 23.02 & 26.78 & 27.46 & 25.07 & 25.67 & 25.36\\
    VILA\cite{lin2024vila} & 6 & LLaMA3-8B & 28.70 & 20.05 & 30.05 & 25.08 & 23.77 & 27.26 & 25.44\\
    LongVA\cite{zhang2024longva} & 32 & Qwen2-Instruct-7B & 27.11 & 23.27 & 26.78 & 29.49 & 27.25 & 25.52 & 26.42\\
    VideoChat2\cite{li2024mvbench} & 16 & Mistral-v0.2-Instruct-7B & 28.93 & 24.75 & 22.95 & 28.47 & 29.13 & 24.09 & 26.72\\
    Valley\cite{luo2023valley} & 8 & LLaMA-13B & 28.70 & 25.00 & 21.86 & \underline{30.51} & 26.38 & 27.73 & 27.02\\
    \toprule
    \multicolumn{10}{c}{\textit{Proprietary}} \\
    \midrule
    GPT-4V\cite{GPT4v} & 8 & - & \underline{36.67} & \textbf{27.72} & \textbf{33.88} & \textbf{32.54} & \textbf{33.62} & \textbf{31.54} & \textbf{32.63}\\
    \bottomrule
\end{tabular}
\end{center}
\end{table*}

\subsection{Experimental Settings}
In this study, we conduct the evaluation on 21 MLLMs, including GPT-4V~\cite{GPT4v}, Video-LLaMA2~\cite{cheng2024videollama}, ShareGPT4Video~\cite{chen2024sharegpt4video}, LLaVA-NeXT-Video~\cite{zhang2024llavanextvideo}, VILA~\cite{lin2024vila}, VideoChat2~\cite{li2024mvbench}, LongVA~\cite{zhang2024longva}, Video-LLaVA~\cite{lin2023video}, Video-ChatGPT~\cite{maaz2023video}, BLIP-2~\cite{li2023blip}, InstructBLIP~\cite{dai2023instructblip}, InstructBLIP Vicuna~\cite{dai2023instructblip}, Yi-VL~\cite{young2024yi}, Qwen-VL-Chat~\cite{bai2023qwen}, Valley~\cite{luo2023valley}, DeepSeek-VL~\cite{lu2024deepseek}, LLaVA1.5~\cite{liu2024improved}, mPLUG-Owl-2~\cite{ye2024mplug}, MultiModal-GPT~\cite{gong2023multimodal}, InternVL-1.5~\cite{chen2024internvl,chen2024far} and InternVL-2~\cite{chen2024internvl,chen2024far}. For video MLLMs, we adhere to their official configurations, including the number of frames. We crop the task progress video as the visual input and modify the sampling function to ensure inclusion of both the first frame and the last frame (representing the current observation image). For image MLLMs, we consistently use 8 key frames, reducing the number if necessary to prevent inference issues. These key frames are uniformly sampled from the provided video clips and saved in advance for model evaluation.

For the evaluation, we use a common prompt as: \textit{Select the best answer to the following multiple-choice question based on the video. Respond with only the letter (A, B, C, or D) of the correct option. Considering the progress shown in the video and my current observation in the last frame, what action should I take next in order to [task goal]? [candidate choices]}. Following the evaluation strategy in Video-MME~\cite{fu2024video}, the accuracy is calculated by matching the output of the model with the real one, without introducing any third party model such as GPT.

\subsection{Main Evaluation Results}
\label{sec:main_results}
The main evaluation results are presented in the last column of Tab.~\ref{tab:main_results}. Considering the test samples are multiple-choice questions with 4 candidate options, the accuracy of random guessing stands at 25\%. However, our observations indicate that most MLLMs struggle at the level of random guessing (23\%-27\%) and fail to demonstrate effective task planning capabilities. Only a few MLLMs, such as Qwen-VL-Chat, InternVL-2 and GPT-4V, achieve total accuracies around 30\%, with the best performer GPT-4V reaching only 32.63\%. 

We summarize three dominant sources that contributed to the challenges posed by EgoPlan-Bench2. First, \textbf{the current observation images} exhibit complex and diverse scenes, characterized by dynamic camera transitions and perspective shifts. These images include a wide array of objects with varying sizes and cluttered backgrounds. Accurately identifying human-object interactions and discerning the states of manipulated objects in such settings is difficult. Second, \textbf{the historical task progress videos} demand an identifying of fine-grained actions and precise comprehension of task progress. This places high requirements on the model’s temporal reasoning and visual perception capabilities. Third, \textbf{the integrated planning process} requires synthesizing information from both the task progress and the current observation. The wide range of scenarios in EgoPlan-Bench2, including tasks from specialized domains, further necessitates substantial world knowledge for informed decision-making. 

The main results underscore a significant gap in achieving human-level task planning capabilities among existing MLLMs, highlighting substantial areas for improvement. We will analyze them in detail in Sec.~\ref{section: case study} based on the above three dominant challenge sources.

\subsection{Study on Different Domains and Scenarios}

\begin{figure*}[ht]
    \centering
    \includegraphics[width=1\textwidth]{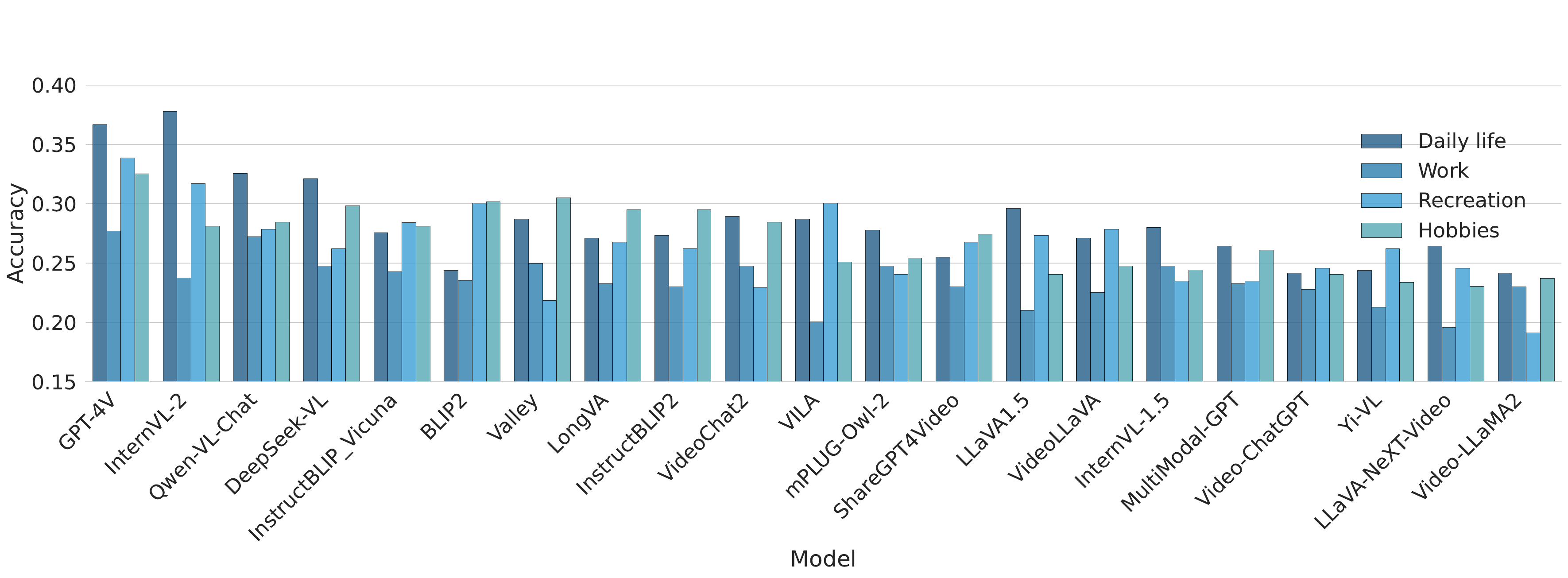}
    \caption{The accuracy of 21 MLLMs in different domains of EgoPlan-Bench2. Most MLLMs demonstrate superior performance in questions associated with Daily life, while exhibiting diminished effectiveness in Work-related questions. Various models display distinct behaviors when addressing issues within domains of Recreation and Hobbies.}
    \label{fig:results_domains}
\end{figure*}

\begin{figure}[h!]
    \centering
    \includegraphics[width=0.48\textwidth]{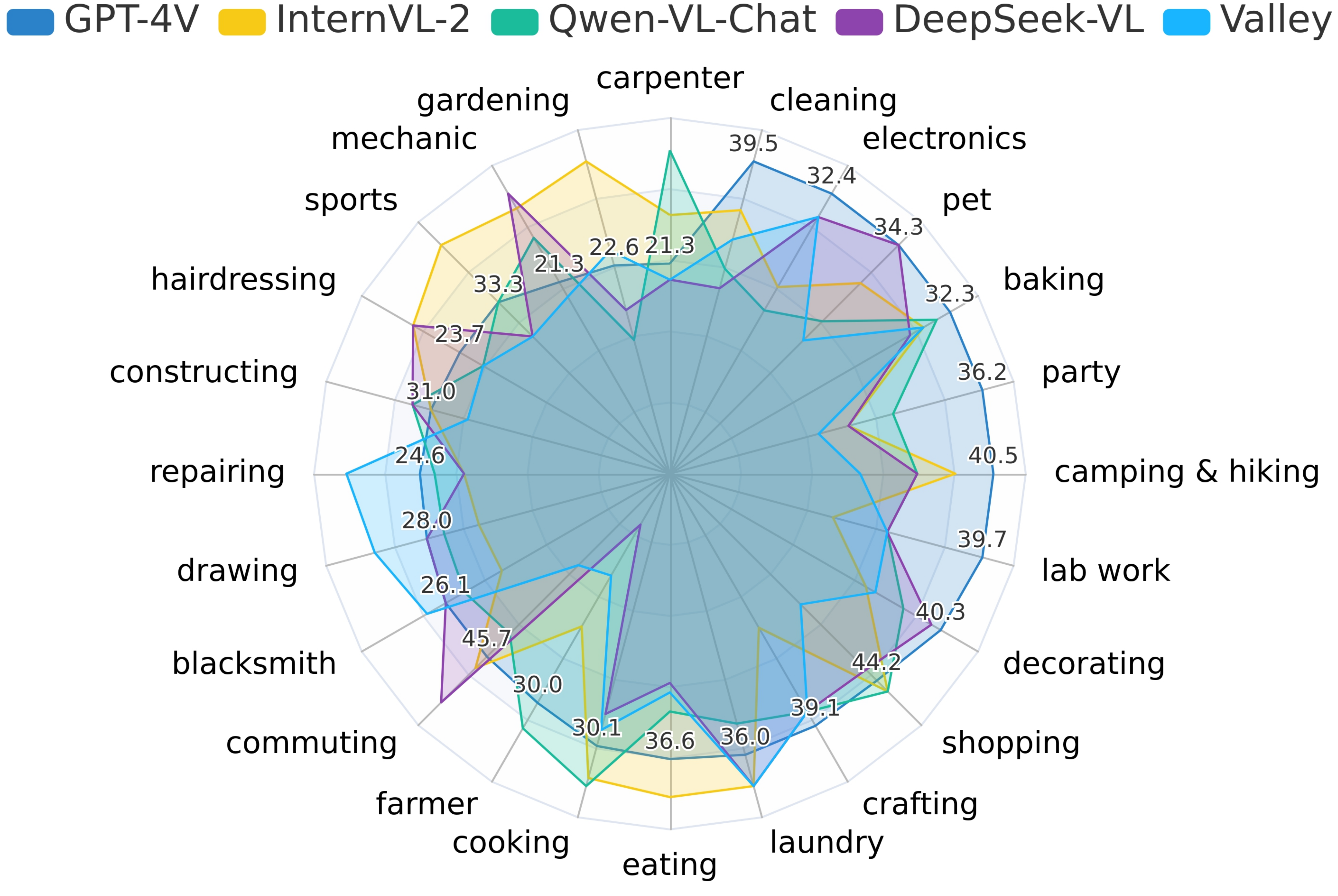}
    \caption{The accuracy of the top-5 performing MLLMs, which include GPT-4V, InternVL-2, Qwen-VL-Chat, DeepSeek-VL and Valley, across the 24 fine-grained scenarios in EgoPlan-Bench2. GPT-4V and InternVL-2 lead at most scenarios and achieve the best overall performance.}
    \label{fig:results_radar}
\end{figure}

The primary attribute of our proposed EgoPlan-Bench2 lies in its authentic and diverse real-world scenarios, emphasizing the assessment of MLLMs' performance across various scenario settings. Tab.~\ref{tab:main_results} and Fig.~\ref{fig:results_domains} illustrate the evaluation results across various domains, while the detailed results for the top-5 performing MLLMs across 24 scenarios are depicted in Fig.~\ref{fig:results_radar}. GPT-4V emerges as the most capable MLLM across three major domains, recording accuracies of 27.72\% in Work, 33.88\% in Recreation, and 32.54\% in Hobbies. In the Daily Life domain, GPT-4V’s performance reaches 36.67\%, closely following InternVL-2, which leads with 37.81\%. 

We conduct an analysis on the impact of different domains on the planning performance of three models (Qwen-VL-Chat, InternVL-2 and GPT-4V) that significantly outperform random guessing. We observe that MLLMs perform optimally in scenarios pertaining to Daily Life and least effectively in Work-related scenarios. Daily Life scenarios, which include tasks like cleaning, laundry and cooking, generally require only basic life experience, readily available in internet-scale datasets and more familiar to MLLMs. Conversely, Work-related scenarios such as those in laboratories, blacksmithing, or carpentry demand complex expertise and familiarity with technological processes and uncommon tools (e.g., lubricants, goggles and sterilizer machines), posing substantial challenges to MLLMs in planning tasks.

\begin{figure*}[h!]
    \centering
    \includegraphics[width=1\textwidth]{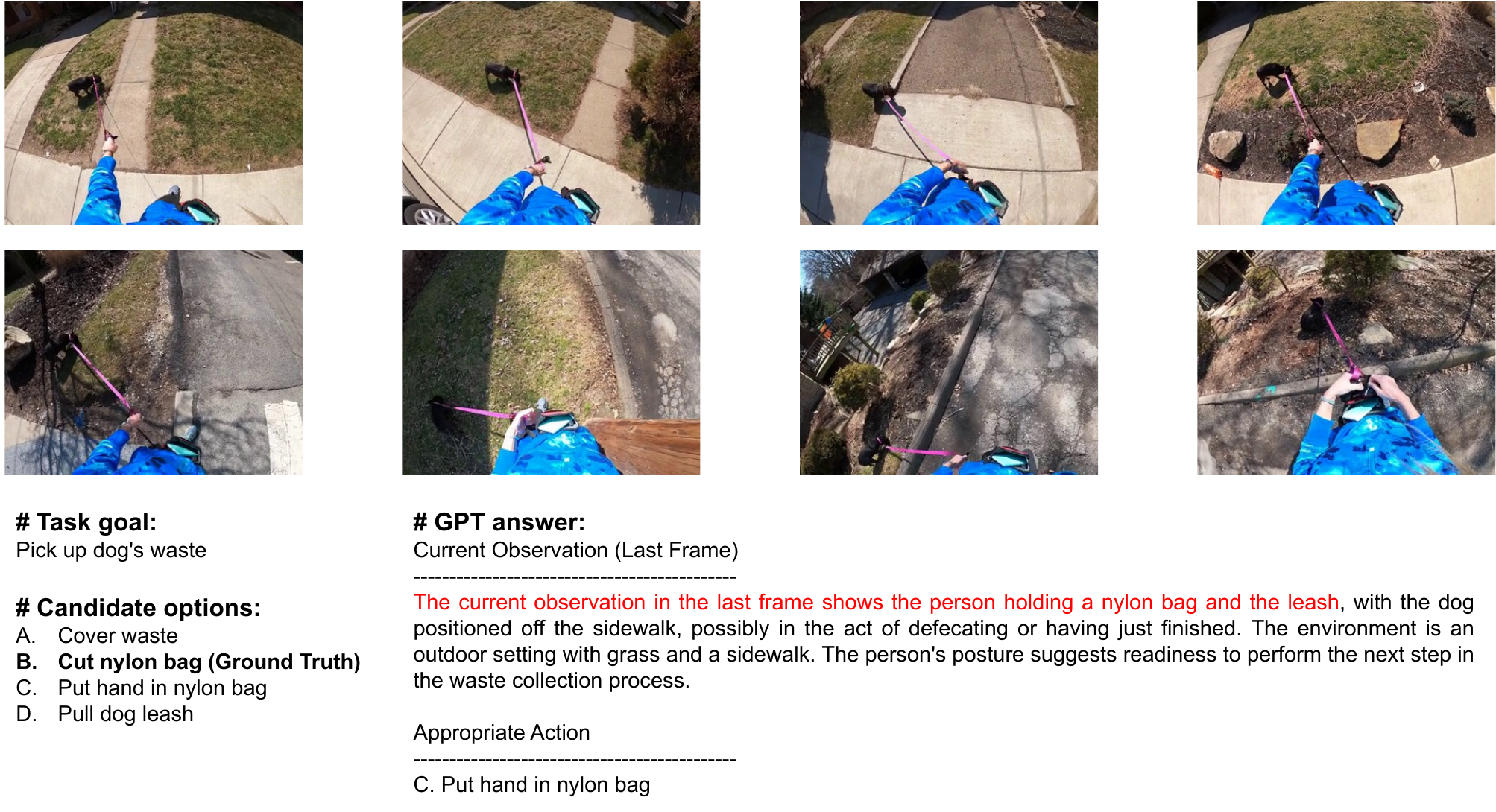}
    \caption{A failure case caused by the \textbf{misperception of the current state (Type I)}. Eight key frames are uniformly sampled in sequence from the video clip, with the final image as the current observation state. The historical task progress, illustrated in the first seven images, shows the camera wearer walking a dog. In the current observation state, she stopped to roll out a trash bag, with the next action involving cutting the nylon bag to collect the dog’s waste. While GPT-4V can provide a general description of the current observation image, it fails to capture the detailed state of the nylon bag as it is being rolled out.}
    \label{fig:case_observation}
\end{figure*}
\begin{figure*}[h!]
    \includegraphics[width=1\textwidth]{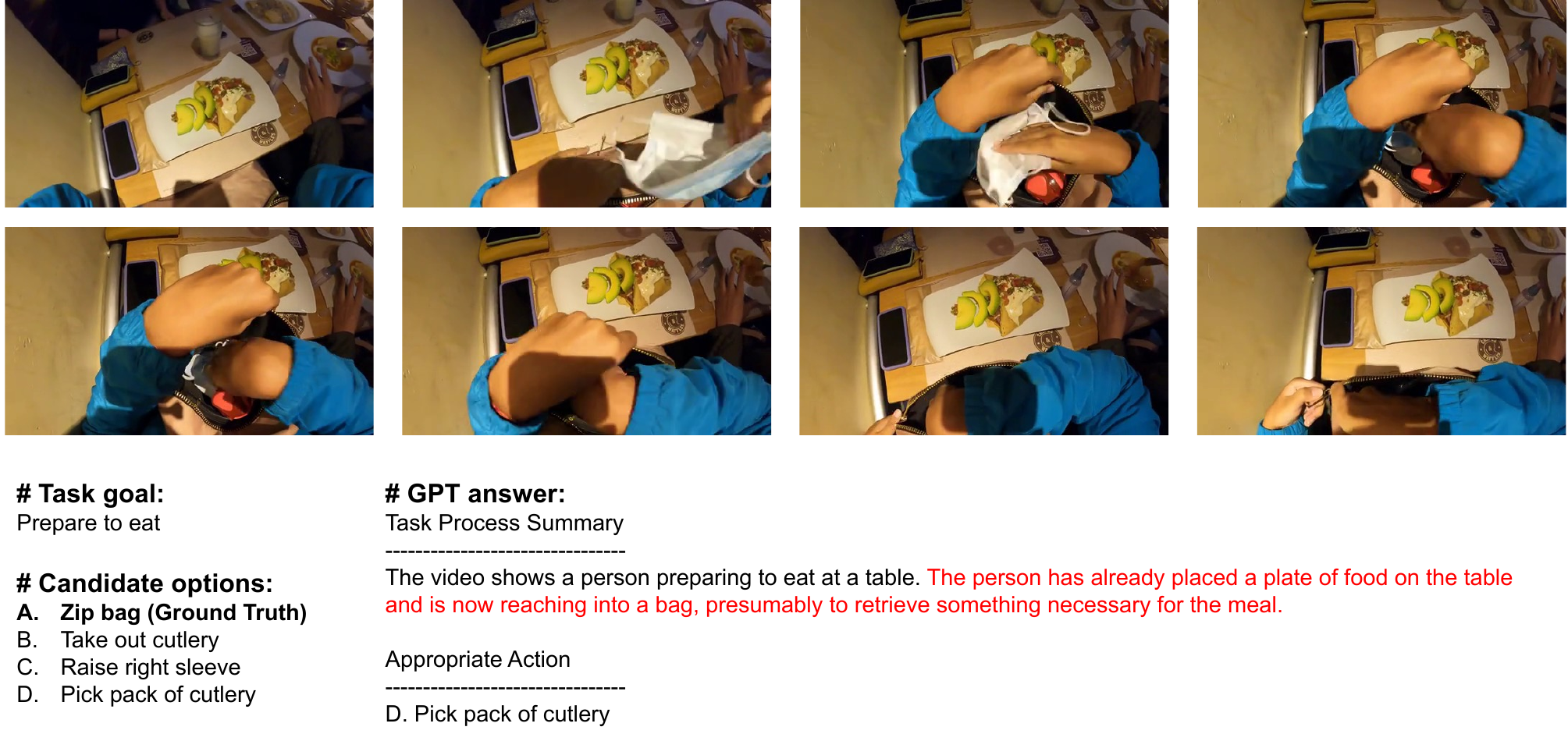}
    \caption{A failure case caused by the \textbf{misunderstanding of the task progress (Type II)}. In the historical task progress, the camera wearer places a mask into a bag and prepares to zip it before retrieving cutlery from the table. However, GPT-4V misinterprets the video clip, erroneously assuming that the camera wearer is reaching into the bag to find something. This misunderstanding leads to an incorrect answer to take pack of cutlery from the bag.}
    \label{fig:case_taskprogress}
\end{figure*}
\begin{figure*}[h!]
    \includegraphics[width=1\textwidth]{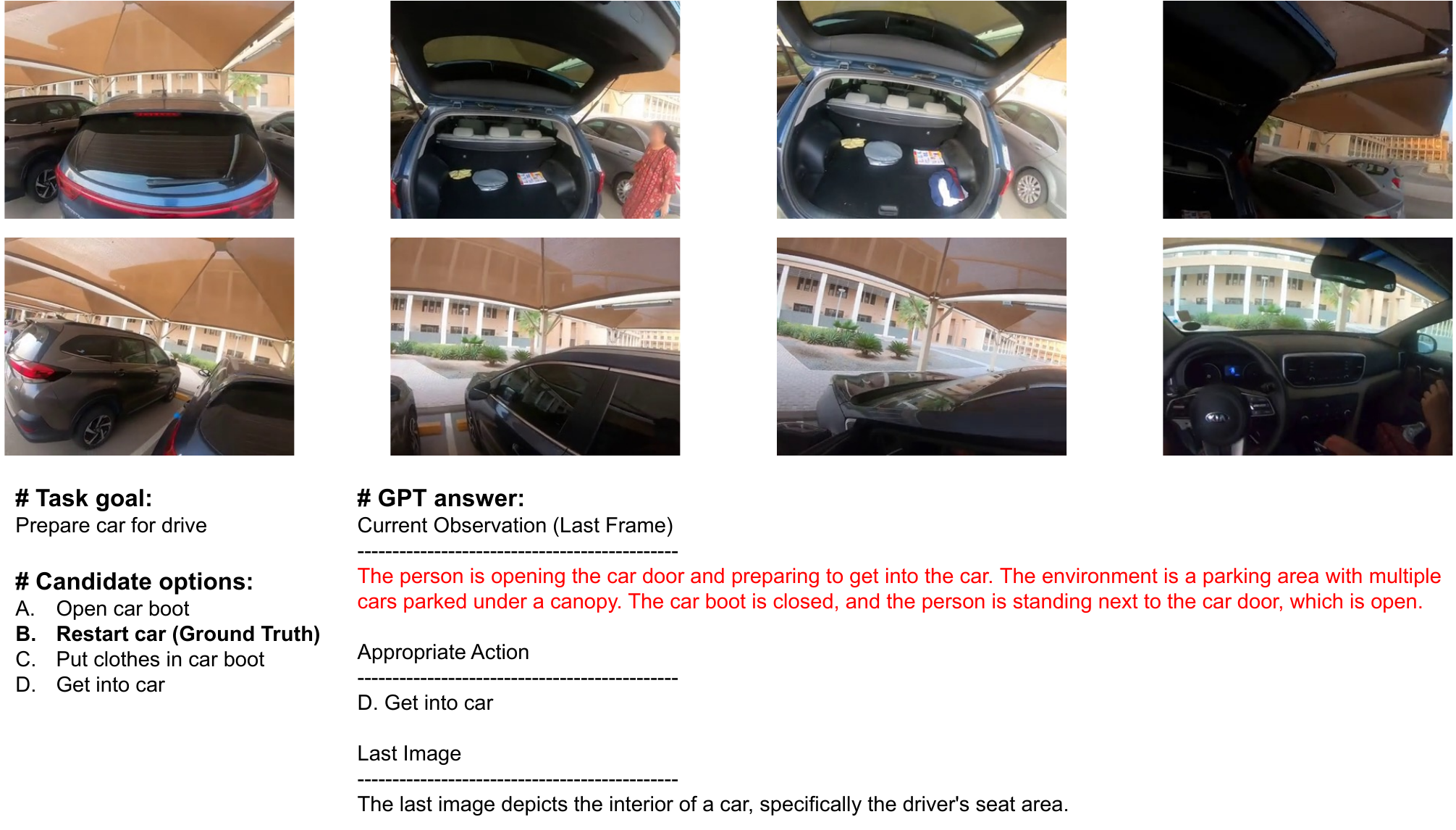}
    \caption{A failure case caused by the \textbf{lack of temporal perception and cognition (Type III)}. In the historical task progress, the camera wearer organizes items in the car boot and opens the car door. He then enters the car and takes out the key to restart the car. When GPT-4V is queried solely about the content of the last image, it accurately identifies that the camera wearer is inside the car. However, when prompted about the current state, it incorrectly conflates the historical task progress with the present moment, mistakenly assuming that the camera wearer is outside the car, preparing to enter.}
    \label{fig:case_temporal}
\end{figure*}
\begin{figure*}[h!]
    \includegraphics[width=\textwidth]{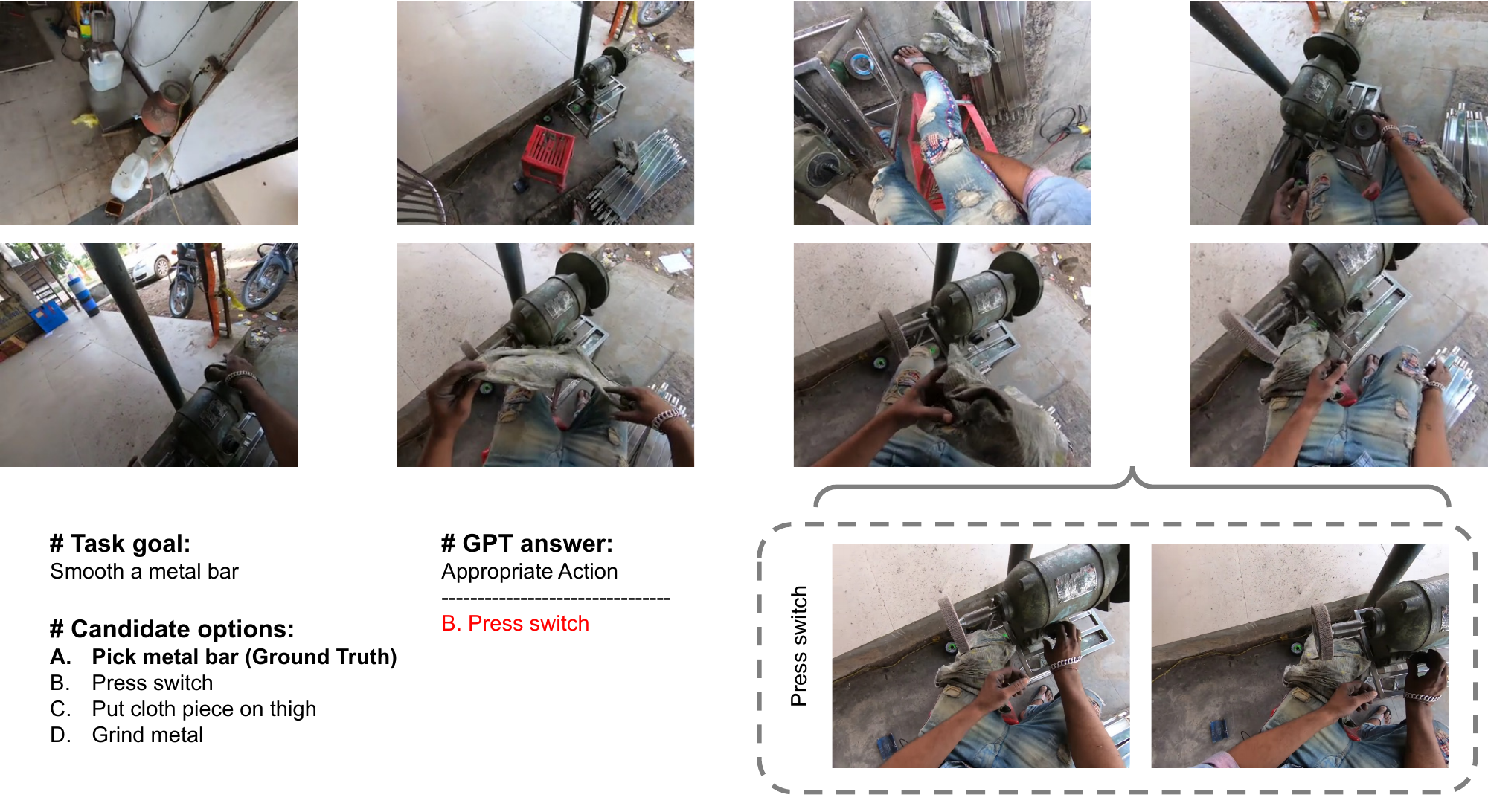}
    \caption{A failure case caused by the \textbf{limitation on the number of sampled frames (Type IV)}.
    In the given video clip, the camera wearer places a cloth piece on the thigh and then turns on a machine, preparing to pick up a metal bar for smoothing. Due to the brief nature of the machine activation, this action is omitted in the uniformly sampled eight key frames. The action of turning on the machine becomes observable only through denser sampling between the seventh and eighth frames, as indicated in the gray box.}
    \label{fig:case_keyframes}
\end{figure*}
\begin{figure*}[h!]
    \includegraphics[width=\textwidth]{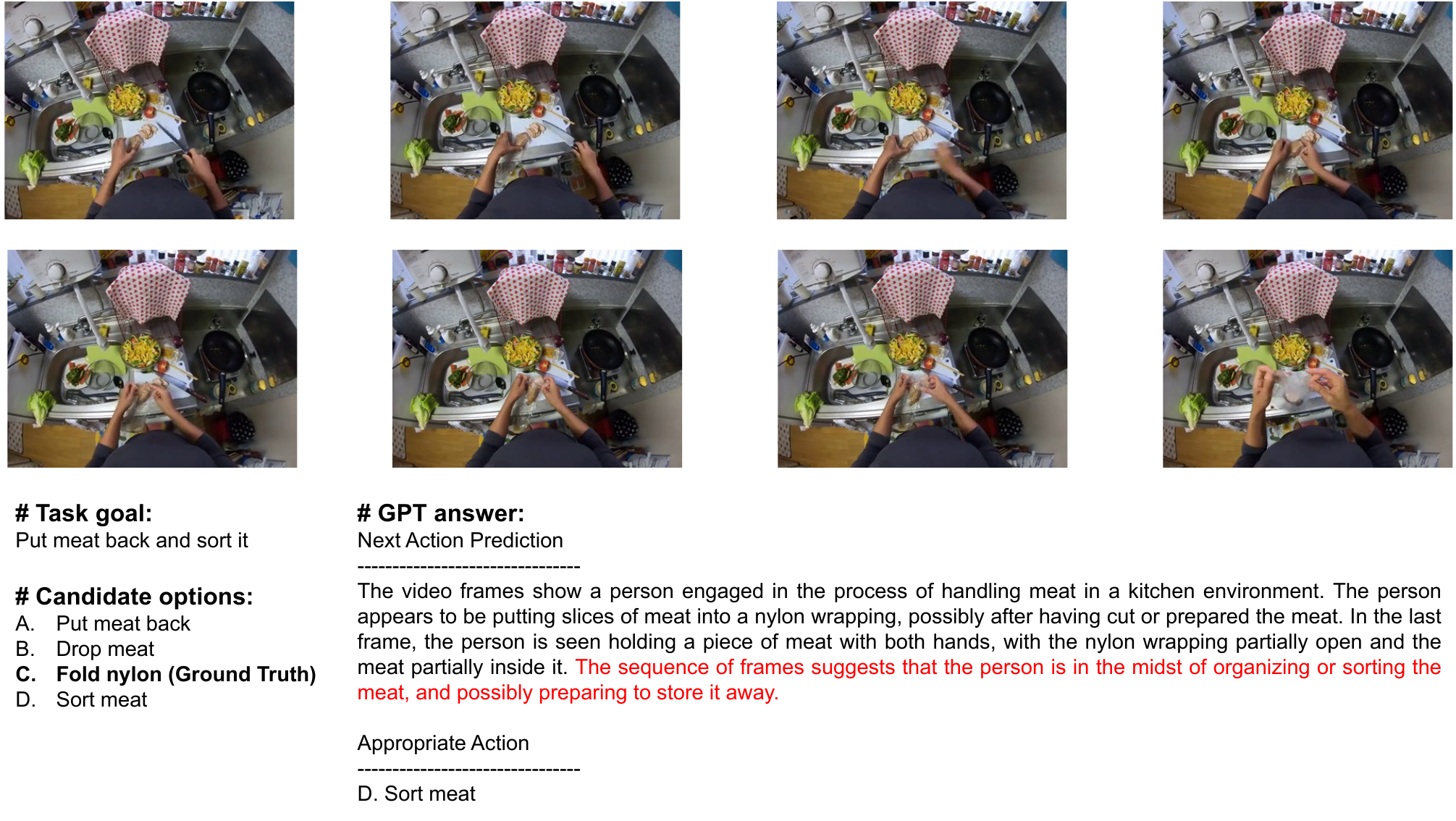}
    \caption{A failure case caused by the \textbf{poor reasoning capacity (Type V)}. In the historical task progress, the camera wearer cuts the meat and places it in a nylon bag. In the current observation image, he picks up an unfolded nylon bag and prepares to fold it. Although GPT-4V accurately describes the task progress and the current observation state, it fails to deduce that the nylon bag needs to be folded before the meat can be stored away.}
    \label{fig:case_reasoning}
\end{figure*}

\subsection{Study on Varying Video Lengths}
The diversity of EgoPlan-Bench2 is also reflected in the variation of video durations. The penultimate and antepenultimate columns in Tab.~\ref{tab:main_results} present the performance of MLLMs across varying video durations. Under the division by video length, GPT-4V achieves superior performance in both short and long video categories, with accuracy scores of 33.62\% and 31.54\%, respectively. The results reveal a decline in MLLM performance as video length increases. This may attributed to the fact that the evaluated MLLMs sample a constant number of frames for videos of varying lengths, despite longer videos typically involving more complex tasks. This may compromise the models' ability to track critical information in long-horizon task progress, which is essential for planning subsequent actions.

\subsection{Error Analysis}
\label{section: case study}
To investigate why MLLMs fall short of expected performance in planning tasks, we conduct an in-depth case study using GPT-4V. According to the three dominant sources of challenges analyzed in Sec.~\ref{sec:main_results}, we identify the following five primary failure types: Type I arises from the current observation state; Types II, III, and IV stem from the historical task progress; Type V originates from the integrated planning process.

\subsubsection{Type I: Misperception of Current State}
For effective planning, MLLMs need to accurately identify details about human-object interactions and the surrounding environment from the current observation image. Nevertheless, MLLMs frequently miss or misidentify objects and overlook detailed interactions between human and objects. As displayed in Fig.~\ref{fig:case_observation}, while the historical task progress shows the camera wearer walking the dog and preparing to pick up its waste, the current observation image reveals a rolled-out nylon bag, indicating that the next action should involve cutting the nylon bag. However, GPT-4V only recognizes that the trash bag is being held, without noticing the detail that it has been rolled out, and incorrectly predicts the next action to be placing the hand on the trash bag.

\subsubsection{Type II: Misunderstanding of Task Progress}
MLLMs demonstrate limited understanding of the historical task progress depicted in videos. While they can grasp the rough scene, they frequently overlook crucial actions and specific details. For example, in Fig.~\ref{fig:case_taskprogress}, the camera wearer is shown placing a mask in a bag before eating, suggesting that the subsequent step should be to zip the bag. However, GPT-4V only perceives the camera wearer reaching into the bag, without noticing the mask placement. So it erroneously predicts that the camera wearer is retrieving cutlery, which is actually on the table.

\subsubsection{Type III: Lack of Temporal Perception and Cognition}
MLLMs’ inability to perceive and cognitively process temporal information results in confusion over the progress of tasks and action sequences at different moments. As shown in Fig.~\ref{fig:case_temporal}, historical video footage shows the camera wearer organizing items in the car boot and then opening the car door. In the current observation, the camera wearer is inside the car, taking out the key, poised to start the vehicle. Although GPT-4V accurately understands the scenes and actions, it misjudges their chronological sequence, mistakenly interpreting images from earlier as part of the current state.

\subsubsection{Type IV: Limitation on the Number of Sampled Frames}
The majority of existing MLLMs are constrained by the number of video frames they can process. This is particularly problematic in long videos where sparse and uniform sampling of video frames may lead to the omission of critical information, neglecting completed actions or changes in states. For instance in Fig.~\ref{fig:case_keyframes}, when the camera wearer puts a cloth piece on the thigh and press the switch of the machine, it indicates that the subsequent step involves picking up a metal bar for smoothing. However, due to the brief duration of the switch activation, key frames captured fail to include this pivotal action, and GPT-4V selects the wrong action which has already been completed.

\subsubsection{Type V: Poor Reasoning Capacity}
Effective task planning requires MLLMs to leverage basic human world knowledge for appropriate reasoning. While MLLMs need to comprehend task goals expressed in language and know about relevant objects and tools, they also must grasp general task processes to make informed inferences about the subsequent action. As shown in Fig.~\ref{fig:case_reasoning}, though GPT-4V correctly describes both the historical task process and the current observation state where the camera wearer picks up an unfolded nylon bag, it fails to deduce that the bag must be folded before the meat is stored to better preserve it.

\subsubsection{Future Directions}
Based on the error analysis which reveals several critical limitations of existing MLLMs, we further discuss some potential avenues for future improvement:
\begin{itemize}
    \item \textbf{Visual Perception and Cognition}: Enhancing visual perception and cognition is vital for accurately interpreting task progress and the current observation state, thereby preventing failure cases such as Type I and Type II. Enhancements should focus on developing the ability to recognize and understand detailed information, including critical actions and interactions between humans and objects.
    \item \textbf{Complex Temporal Understanding}: Existing MLLMs exhibit limited temporal understanding, which hampers their ability to comprehend the sequence of actions effectively, as evidenced by Type III failures. Future developments should concentrate on advancing temporal reasoning capabilities, enabling MLLMs to accurately interpret the chronological order of events and make more coherent and logical action predictions.
    \item \textbf{Long Context Modeling}: Type IV failures reveal that restricted input frames often serve as an information bottleneck, limiting the understanding of video content in its entirety. MLLMs should process an increased number of input frames to mitigate information loss when sampling from lengthy videos.
    \item \textbf{Reasoning Ability}: Strengthening the reasoning abilities of MLLMs is essential for effective task planning. As shown in the failure case of Type V, models should incorporate both foundational and specialized knowledge of the human world, enabling them to understand basic task workflows and apply this knowledge in conjunction with the specific context of current tasks.
\end{itemize}

\section{Towards Human-level Planning with Multimodal Chain-of-Thought Prompting}


In the realm of natural language processing, Chain-of-Thought (CoT) reasoning empowers language models to tackle complex tasks by informing them to generate intermediate rationales. Numerous recent studies~\cite{wang2022language}, ~\cite{jiang2024joint}, ~\cite{wang2024vlm}, ~\cite{mitra2024compositional}, ~\cite{zhou2024image} have significantly enhanced the performance of MLLMs through the application of multimodal CoT prompting, which integrates CoT reasoning with additional multimodal prompts. 

\begin{figure*}[h!]
    \includegraphics[width=\textwidth]{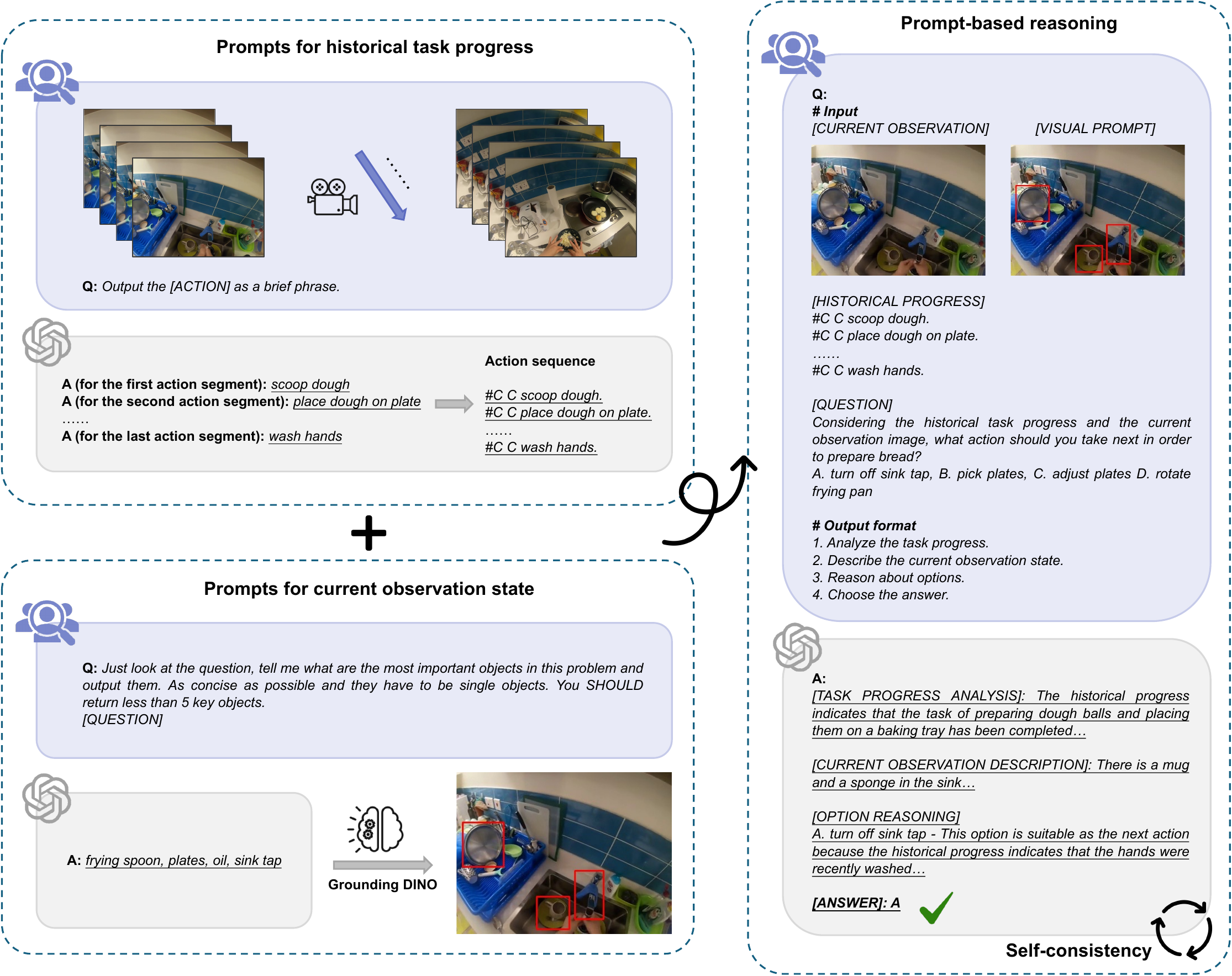}
    \caption{The pipeline of our training-free multimodal Chain-of-Thought (CoT) prompting method. We utilize predicted actions sequences as a prompt for representing historical task progress, and bounding box of key objects as a prompt to enhance the understanding of visual observations. By combining these elements with CoT reasoning and a self-consistency mechanism, we strengthen GPT-4V's planning capabilities without the need for additional training.}
    \label{fig:cot}
\end{figure*}

When evaluated on EgoPlan-Bench2, MLLMs encounter challenges in making direct planning decisions. In this section, we propose a flexible and effective multimodal CoT prompting approach aimed at improving model performance and analyzing their bottlenecks in task planning. In Sec.~\ref{sec:main_results}, we discussed three dominant sources of challenges in EgoPlan-Bench2: two types of visual information including current observation image and historical task progress video, and the integrated planning process. Our objective is to improve the model's planning capabilities through multimodal CoT prompting without additional training, focusing on these three aspects. Specifically, we begin with a preliminary study addressing performance bottlenecks through auxiliary multimodal prompts tailored to those two types of visual information. 
For the integrated planning process, we introduce a prompt-based reasoning strategy. We employ CoT reasoning with GPT-4V to generate step-by-step rationales, facilitating integration of multimodal input alongside auxiliary prompts. Finally, a multi-iteration decision approach is applied to reinforce answer consistency. The pipeline is depicted in Fig.~\ref{fig:cot}. Please refer to supplementary materials for detailed multimodal prompts.

\begin{table*}
\begin{center}
\caption{The impact of different types of multimodal prompts and prompt-based reasoning strategies on the performance of GPT-4V.}
\label{tab:prompts}
\begin{tabular}{cccc}
    \toprule
    Task progress prompts & Current observation state prompts & Prompt-based reasoning & Acc \\
    \midrule
    \midrule
    - & - & - & 32.80\\
    \midrule
    Action-seq-GT & - & - & 51.67(+18.87)\\
    Action-seq-GPT & - & - & 36.71(+3.91)\\
    Description-video & - & - & 32.45(-0.35)\\
    Description-frame & - & - & 32.91(+0.11)\\
    Tracking & - & - & 31.88(-0.92)\\
    \midrule
    - & Description-img & - & 33.83(+1.03) \\
    - & BoundingBox-hand & - & 36.82(+4.02) \\
    - & BoundingBox-obj & - & 37.63(+4.83) \\
    - & Status-obj & - & 29.46(-3.34) \\
    - & SG (only obj) & - & 32.57(-0.23) \\
    - & SG (obj + attribute) & - & 31.99(-0.81) \\
    - & SG (obj + relation) & - & 31.53(-1.27) \\
    - & SG (obj + attribute + relation) & - & 31.88(-0.92) \\
    \midrule
    - & BoundingBox-obj & CoT & 39.82(+7.02)\\
    Action-seq-GPT & BoundingBox-obj & CoT & 42.81(+10.01) \\
    Action-seq-GPT & BoundingBox-hand & CoT & 40.97(+8.17) \\
    Action-seq-GPT & BoundingBox-hand \& obj & CoT & 41.77(+8.97) \\
    Action-seq-GPT & BoundingBox-obj & CoT \& Self-refinement & 42.46(+9.66) \\
    Action-seq-GPT & BoundingBox-obj & CoT \& Self-consistency & \textbf{43.04(+10.24)} \\
    \bottomrule
\end{tabular}
\end{center}
\end{table*}

\subsection{Prompts for Historical Task Progress}
In EgoPlan-Bench2, historical task progress is presented in an egocentric video format, with durations ranging from a few seconds to five minutes. Previous experiments detailed in Sec.~\ref{section: case study} reveal that while MLLMs can comprehend the general scene depicted in the video, they are inclined to overlook crucial actions and demonstrate poor temporal perception and cognition. To mitigate the deficiencies in historical task progress understanding, we employ the following prompts:
\begin{itemize}
    \item \textbf{Action sequence} (Action-seq).
    Following the long-term memory extraction method described in ~\cite{shi2024epd}, we generate a text sequence of actions (Action-seq-GPT) that captures the historical task progress by GPT-4V, and also extract raw annotations to construct a ground-truth action sequence (Action-seq-GT) for comparison purposes.
    \item \textbf{Video-level description} (Description-video), a comprehensive description for the input egocentric video clip.
    \item \textbf{Frame-level description} (Description-frame), descriptions of eight key frames, which are concatenated in sequential order to provide a frame-by-frame account of the video content.
    \item \textbf{Key objects tracking} (Tracking). The motion trajectories of key objects often contain important task information indicating their future movement patterns, which may assist in planning the subsequent action. Inspired by ~\cite{wang2024vlm}, we utilize Grounding DINO~\cite{liu2023grounding} and SAM2 ~\cite{ravi2024sam} to facilitate key object tracking.
\end{itemize}

The results are shown in the second part of Tab.~\ref{tab:prompts}. Due to some videos triggering Azure OpenAI’s content filtering policy, we retain a total of 869 QA pairs for analysis. All experimental results shown in Tab.~\ref{tab:prompts} are based on these 869 QA pairs. Compared to the case without any additional prompts, summarizing historical videos into concise action sequences markedly improves GPT-4V’s planning ability. However, employing descriptions of videos and video frames, as well as tracking key objects, does not produce a noticeable effect. The action sequences summarized by GPT-4V exhibit a high degree of temporal structuring, facilitating the model’s understanding of the overall task flow and the actions completed. In contrast, video and frame descriptions, while providing an overview of the task flow, fail to deliver clear and detailed information on the temporal sequence and often provide overly vague descriptions of actions, missing critical details. Using real action sequences derived from annotations (Action-seq-GT) can further enhance the accuracy of planning tasks. This finding underscores the importance of precise and temporally structured action sequence information in historical egocentric videos, even it brief, over more elaborate scene descriptions and motion trajectories of key objects.

\subsection{Prompts for Current Observation State}
Another critical visual cue is the current observation image, which reflects the spatial relationships, interactions, and statuses between the camera wearer and manipulated objects. This information can impact the rationality of the subsequent action and the execution success rate, thus offering valuable insights for planning. In this section, we analyze the following prompts:
\begin{itemize}
    \item \textbf{Image description} (Description-img), a detailed description of the current observation image.
    \item \textbf{Bounding box of human hands} (BoundingBox-hand). The movements and positions of human hands are indicative of interactions between the camera wearer and the manipulated objects. We delineate bounding boxes around the hands in the current observation image as visual prompt.
    \item \textbf{Bounding box of key objects} (BoundingBox-obj). We also delineate bounding boxes around the key objects, emphasizing interactions between humans and objects and spatial relationships among different objects.
    \item \textbf{Image-based status of key objects} (Status-obj). We crop key objects from the current observation image and concatenate them into an image grid. Compared to the current observation image with bounding boxes, this visual prompt composed of the cropped key objects focuses on the states of the key objects but loses spatial and interaction features.
    \item \textbf{Scene graph} (SG). Scene graph is a formalization of objects and their relations and attributes that has been extensively used as a bridge between the visual and textual domains. We follow the steps of ~\cite{mitra2024compositional} to summarize the current observation image into the text-based scene graph.
\end{itemize}

As shown in the third part of Tab.~\ref{tab:prompts}, the bounding boxes of both human hands and key objects can equip GPT-4V with stronger human-level planning ability by providing the interaction information between humans and manipulated objects. Specifically, the bounding boxes for key objects emphasize the spatial relationships among these critical items. The image-based status of key objects did not exhibit significant effects, indicating that the importance of object status is relatively minor compared to interaction and spatial relationships. Additionally, we find in our experiments that in complex real-world scenarios, the same noun phrases often denote multiple different objects, resulting in Grounding DINO occasionally detecting incorrect target objects. The misidentification of these target objects could detrimentally impact the model’s planning process. Although scene graphs represent the status and spatial relationships of key objects in a textual format, they are not crucial for planning tasks. We attribute this limitation to the coarse granularity of the text format, which fails to provide the fine-grained features available through visual prompts. Moreover, descriptions of current observation images are not task-oriented, leading to inadequate containment of task-relevant detail.

\subsection{Prompt-Based Reasoning Strategy}
Through prior sections, we identify effective prompts for planning tasks, including the action sequence for the historical task progress and two types of bounding boxes for the current observation. In this section, we employ the CoT reasoning approach to facilitate step-by-step task planning and integrate various effective multimodal prompts, as demonstrated in Fig.~\ref{fig:cot}. GPT-4V is instructed to: a) analyze completed actions and historical task progress, b) describe the current observation state based on visual input, c) assess the suitability of options as the next action relative to the task progress and determine their feasibility in the current state, d) choose the best answer from candidate choices. Using only object-related bounding boxes as additional prompts, we compared the results of direct action prediction with those generated through CoT reasoning. Direct prediction achieves an accuracy of 37.63\%, whereas incorporating CoT reasoning improves accuracy to 39.82\%. This demonstrates the importance of generating intermediate reasoning chains in planning tasks.

Finally, we explore two types of multi-iteration decision approaches to reinforce answer consistency. The first involves a \textbf{self-refinement} approach, where GPT-4V iteratively corrects and refines the reasoning steps and answers from previous rounds until it confirms the correctness of the previous response. The second strategy employs a \textbf{self-consistency} mechanism, wherein GPT-4V generates answers for five times and selects the most frequently produced option among multiple answers. By integrating prompts of Action-seq-GPT and BoundingBox-obj with multimodal CoT reasoning and self-consistency, GPT-4V achieves a peak accuracy rate of 43.04\%.

\section{Conclusion}
In this research, we introduce \textbf{EgoPlan-Bench2}, a benchmark specifically designed to evaluate the task planning capabilities of MLLMs across a variety of real-world scenarios. We construct EgoPlan-Bench2 based on three primary principles: the inclusion of diverse real-world scenarios, an egocentric perspective, and a focus on evaluating planning capacity. EgoPlan-Bench2 encompasses everyday tasks spanning four major domains and 24 detailed scenarios that closely reflect human daily life. EgoPlan-Bench2 is developed through a semi-automatic process that utilizes egocentric videos, supplemented by manual verification to ensure accuracy. The evaluation of 21 MLLMs reveals that EgoPlan-Bench2 poses significant challenges to existing models. Using GPT-4V as a case study, we analyze the reasons behind its shortcomings in real-world task planning and provide insights that could guide the future development of MLLMs toward achieving human-level task planning capabilities. To enhance the planning proficiency of current MLLMs, we propose a novel, training-free multimodal Chain-of-Thought (CoT) prompting method. This approach significantly improves the planning performance of GPT-4V by generating intermediate reasoning chains and leveraging various effective prompts.


\bibliographystyle{unsrt}
\bibliography{ref}

\end{document}